\pdfoutput=1

\documentclass[11pt]{article}

\usepackage[preprint]{acl}

\usepackage{times}
\usepackage{latexsym}

\usepackage[T1]{fontenc}

\usepackage[utf8]{inputenc}

\usepackage{microtype}

\usepackage{inconsolata}

\usepackage{graphicx}
\usepackage{courier}
\usepackage{amsmath}
\usepackage{multirow}
\usepackage{multicol}
\usepackage{colortbl}
\usepackage{xcolor}  
\usepackage{subcaption}
\usepackage{float}  
\usepackage{afterpage}
\usepackage{caption}
\usepackage{array}
\usepackage{makecell}
\usepackage{placeins}  
\usepackage{afterpage}  

\usepackage[most]{tcolorbox}
\usepackage{atbegshi}

\newcommand{\SpiritRAG}{\includegraphics[width=1.5em]{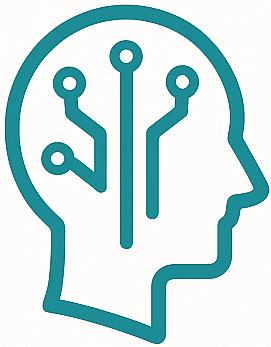}}
\newcommand{\liri}{\includegraphics[width=2.5em]{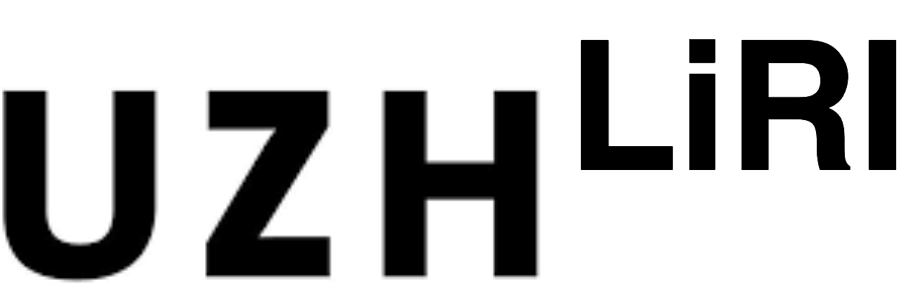}}
\newcommand{\trf}{\includegraphics[width=2.5em]{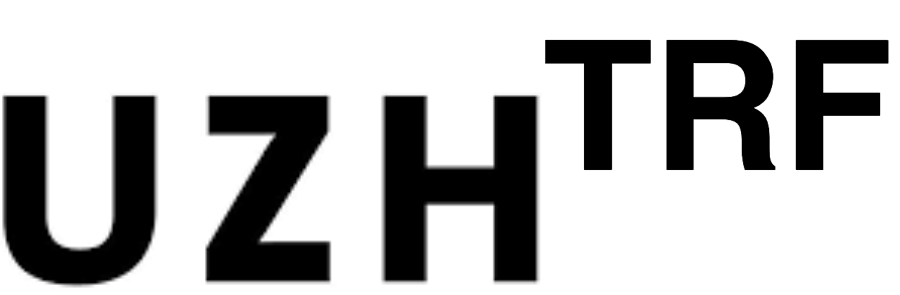}}
\newcommand{\ife}{\includegraphics[width=2.3em]{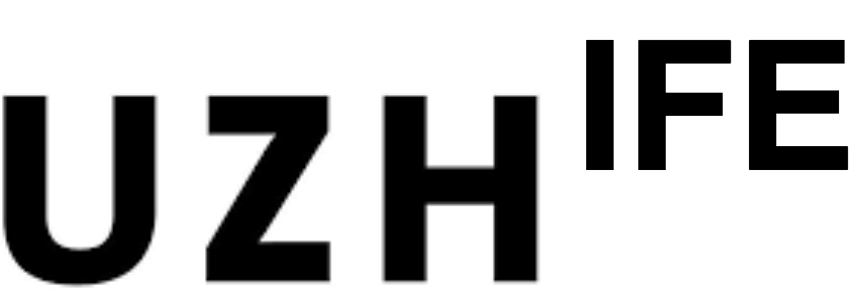}}

\title{
\hspace{-1em}
    \begin{minipage}{2em}
        \centering
        \SpiritRAG
    \end{minipage}
    \hspace{-0.5em}
    \begin{minipage}{0.85\textwidth}
        \centering
        \raisebox{-1.5\height}{
        \shortstack{SpiritRAG: A Q\&A System for Religion and Spirituality in the \\ United Nations Archive
        }
        }
    \end{minipage}
}

\author{
\hspace{1em}
  \textbf{Yingqiang Gao\footnotemark[2]}$^{\liri}$ 
  \textbf{Fabian Winiger\footnotemark[3]}$^{\trf}$ \\
  \hspace{1em}
  \textbf{Patrick Montjourides\footnotemark[3]}$^{\ife}$ 
  \textbf{Anastassia Shaitarova\footnotemark[3]}$^{\liri}$ \\
  \hspace{1em}
  \textbf{Nianlong Gu}$^{\liri}$ 
  \textbf{Simon Peng-Keller}$^{\trf}$
  \textbf{Gerold Schneider}$^{\liri}$ \vspace{0.5em} \\
  \hspace{1em}
  $^{\liri}$Linguistic Research Infrastructure, University of Zurich, Switzerland \\
  \hspace{1em}
  $^{\trf}$Faculty of Theology and the Study of Religion, University of Zurich, Switzerland \\
  \hspace{0.5em}
  $^{\ife}$Institute of Education, University of Zurich, Switzerland \\
  \hspace{0.6em}
  \texttt{yingqiang.gao@uzh.ch} \\
  \url{https://SpiritRAG.linguistik.uzh.ch} 
}

\renewcommand{\thefootnote}{\fnsymbol{footnote}}  

\definecolor{spiritgreen}{HTML}{1C8D96}
\colorlet{lightgreen}{spiritgreen!60} 

\definecolor{all-minilm}{HTML}{2E8B58}
\definecolor{qwen-embedding}{HTML}{077969}

\definecolor{qwen3-tiny2}{HTML}{6856F2}
\definecolor{qwen3-small2}{HTML}{40389E}

\begin{document}

\maketitle
\footnotetext[2]{Corresponding author.}
\footnotetext[3]{Equal contribution.}


\addtolength{\topskip}{4.5em} 

\begin{abstract}

Religion and spirituality (R/S) are complex and highly domain-dependent concepts which have long confounded researchers and policymakers. Due to their context-specificity, R/S are difficult to operationalize in conventional archival search strategies, particularly when datasets are very large, poorly accessible, and marked by information noise. As a result, considerable time investments and specialist knowledge is often needed to extract actionable insights related to R/S from general archival sources, increasing reliance on published literature and manual desk reviews. To address this challenge, we present SpiritRAG, an interactive Question Answering (Q\&A) system based on Retrieval-Augmented Generation (RAG). Built using 7,500 United Nations (UN) resolution documents related to R/S in the domains of health and education, SpiritRAG allows researchers and policymakers to conduct complex, context-sensitive database searches of very large datasets using an easily accessible, chat-based web interface. SpiritRAG is lightweight to deploy and leverages both UN documents and user provided documents as source material. A pilot test and evaluation with domain experts on 100 manually composed questions demonstrates the practical value and usefulness of SpiritRAG.\footnote{SpiritRAG is under MIT license.}

\noindent
\begin{minipage}{\columnwidth}
\centering
\raisebox{-0.15cm}{\hspace{-4em}\includegraphics[height=0.5cm]{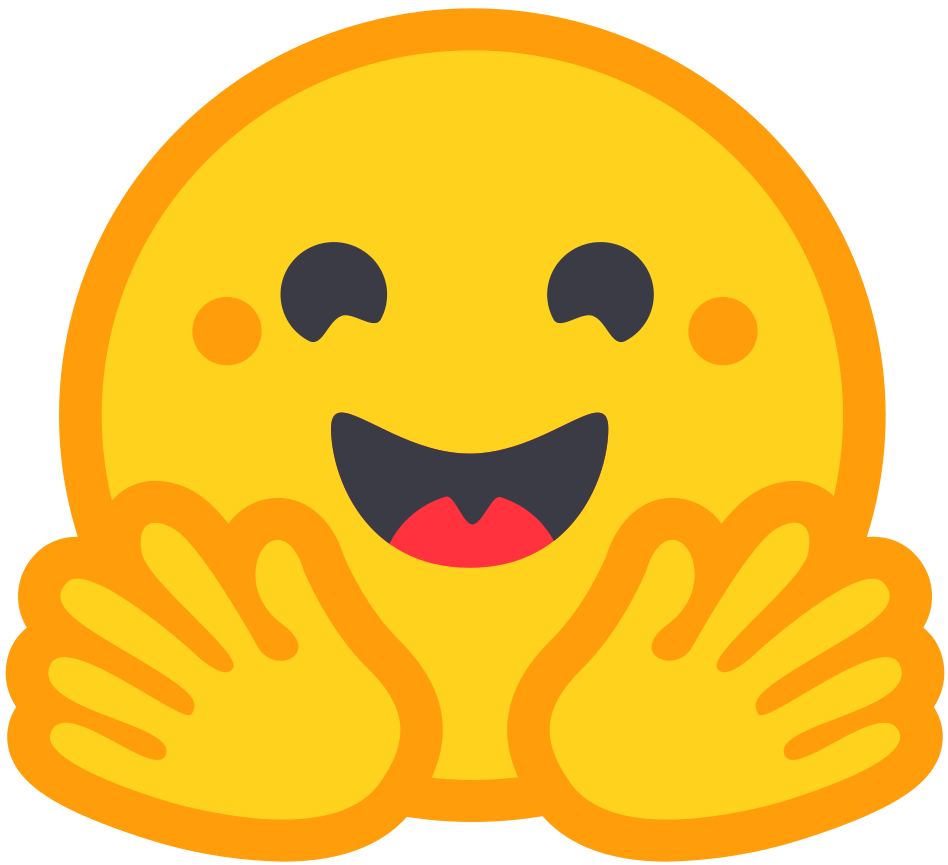}}~
\href{https://huggingface.co/datasets/CharizardAcademy/UN-RES}{\textcolor{gray}{Dataset}} \quad
\raisebox{-0.15cm}{\includegraphics[height=0.5cm]{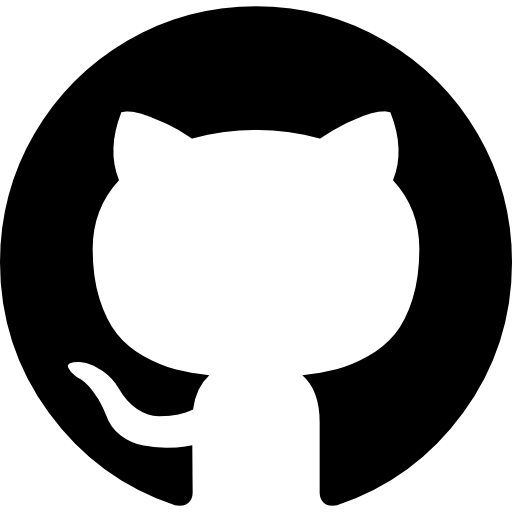}}~
\href{https://github.com/CharizardAcademy/SpiritRAG}{\textcolor{gray}{Code}}
\end{minipage}

\end{abstract}

\renewcommand{\thefootnote}{\arabic{footnote}}

\section{Introduction}

\begin{figure}
    \centering
    \vspace{5em}
    \resizebox{\columnwidth}{!}{
    \includegraphics[width=\columnwidth]{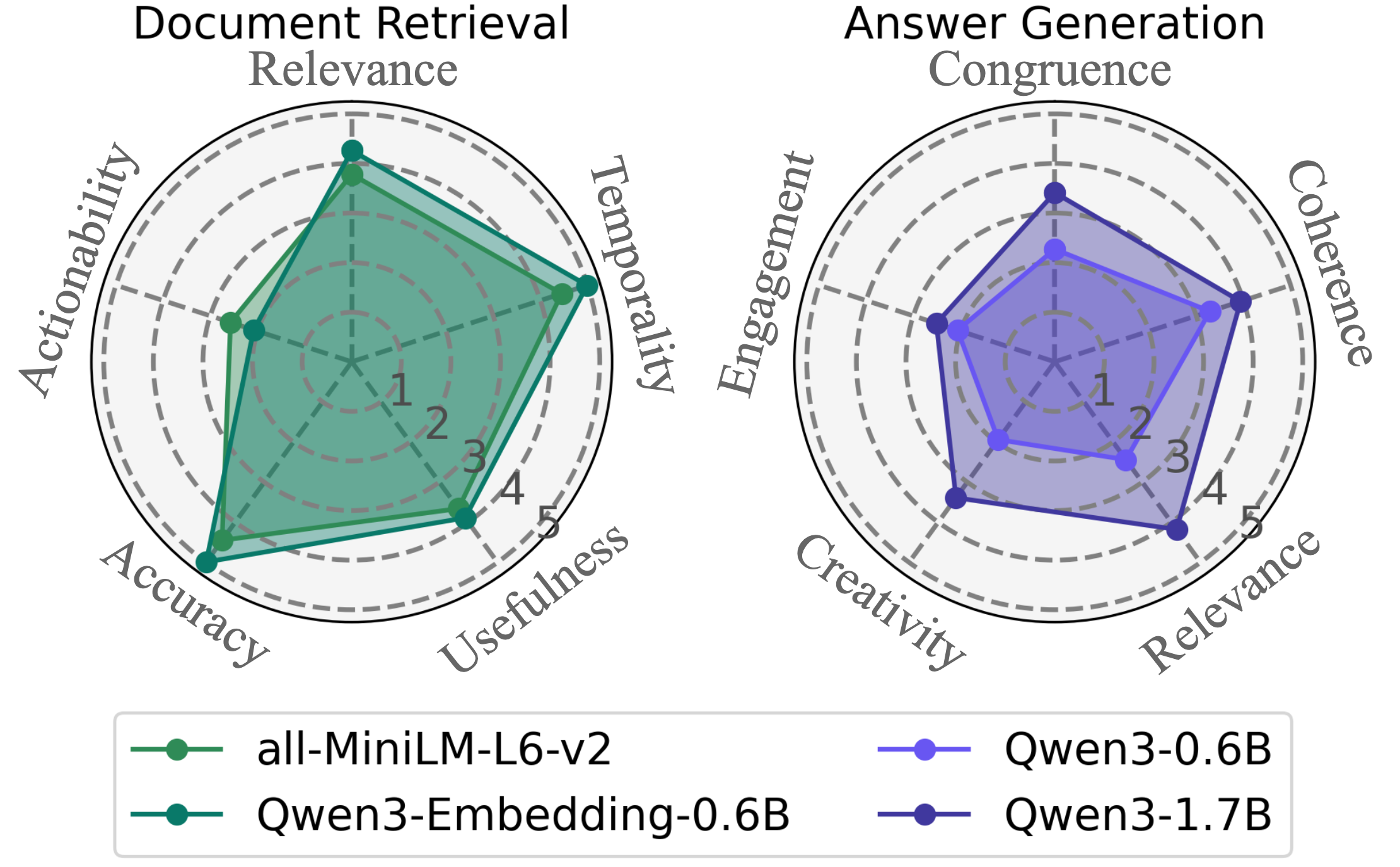}
    }
    \caption{Human evaluation results of SpiritRAG on 50 health + R/S test questions listed in Appendix~\ref{sec:questions}. Our best-performing system setup received promising assessments from domain experts.}
    \label{fig:teaser}
\end{figure}

\setlength{\topskip}{10pt}

The concepts of ``religion'' and ``spirituality'' (R/S) are complex and highly domain-dependent, and vary substantially across social and cultural contexts.  The meaning and connotation of R/S  poses unique challenges to historians, sociologists, anthropologists, and other scholars who seek to understand R/S across distinct social and historical settings. Definitions of R/S vary widely across academic disciplines, and scholars have yet to reach a consensus on how to define them in general terms \citep{steensland2022eminently, winiger2025spiritual}.

For policymakers, the complexity and domain-dependence of these terms are significant barriers to research and decision-making. This is illustrated by the long and protracted history of R/S in the United Nations (UN) system, the main locus of intergovernmental cooperation in the pursuit of global values for humanity, where R/S regularly surfaces as a matter of mutual concern in the areas of healthcare \citep{peng-keller_spirit_2022}, Mental Health and Psychosocial Support (MHPSS) in humanitarian settings \citep{ager_case_2019}, soft diplomacy and value politics \citep{stensvold_religion_2017,  carrette_religion_2017, baumgart-ochse_religious_2019, steiner_religious_2021}, and various development and peace building efforts \citep{un_inter-agency_task_force_on_religion_and_development_engaging_2020}. Accessing and analyzing documents produced by the organizations and agencies of the UN often require extended and time-consuming archival research using poorly designed or outdated search interfaces. This is exemplified by the official archive of the UN\footnote{\url{https://documents.un.org/}}, which includes a vast collection of documents produced by the Security Council, the General Assembly, the Economic and Social Council (ECOSOC) and their subsidiaries, and others. Additionally, some UN entities, such as the World Health Organization (WHO) and the United Nations Educational, Scientific, and Cultural Organization (UNESCO), maintain separate archives.

\begin{figure*}
    \centering
    \includegraphics[width=0.95\textwidth]{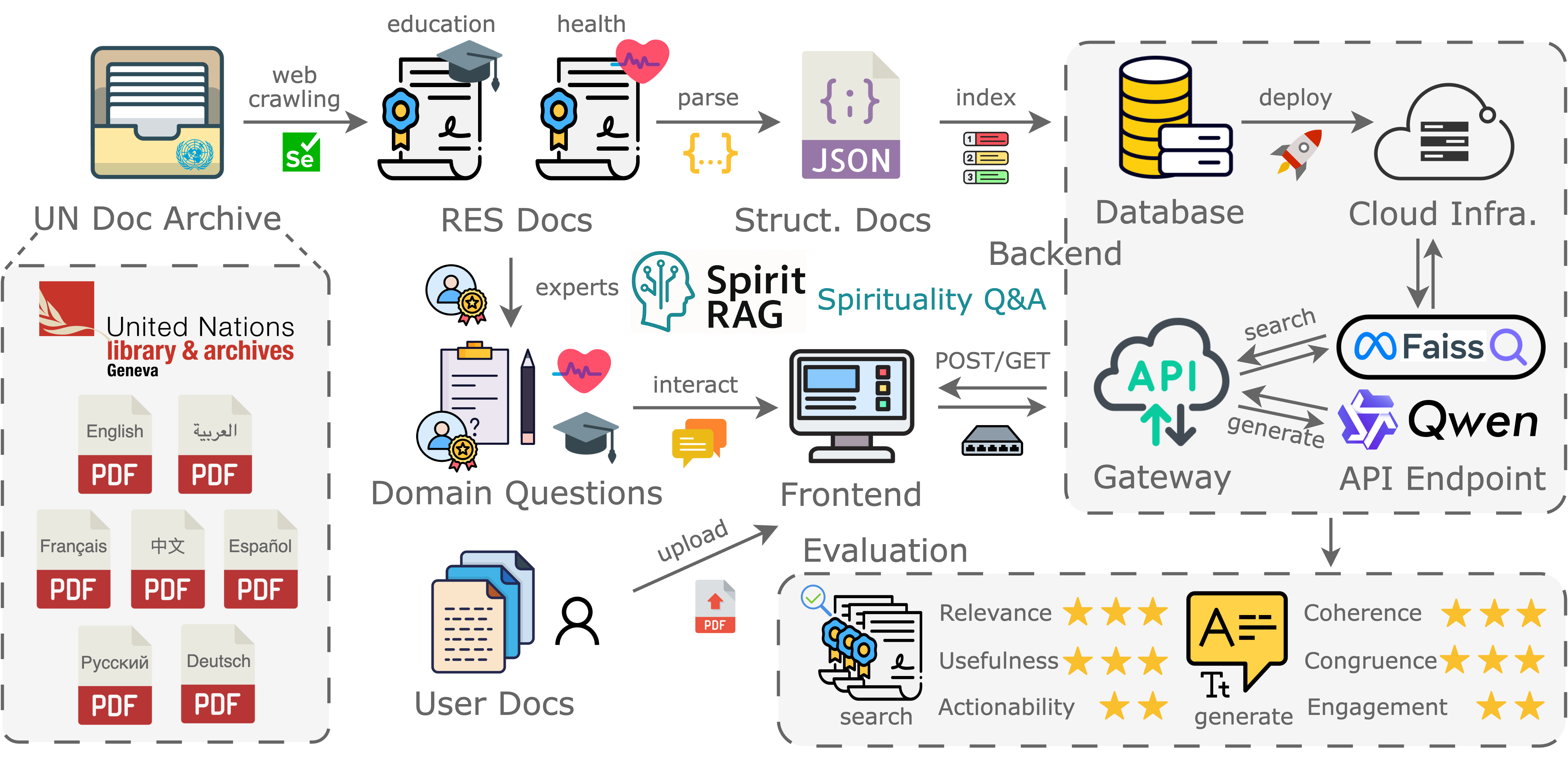}
    \caption{Architecture of SpiritRAG. SpiritRAG is an interactive Question Answering (Q\&A) system enhanced by Retrieval-Augmented Generation (RAG) and specifically designed for acquiring domain-dependent knowledge in the resolution documents (RES Docs) of United Nations (UN) with a focus on Religion and Spirituality (R/S) related to health and education.}
    \label{fig:pipeline}
\end{figure*}


\makeatletter
\setlength{\topskip}{0em}  
\makeatother

The problem of accessibility is greatly exacerbated when searching for information on relatively niche, complex and domain-dependent concepts such as R/S, in which case the retrieval of actionable information requires a substantial investment of time to manually search very large numbers of often largely irrelevant documents marked by information ``noise'' such as procedural formalities, preambular clauses, extensive appendices, and frequently used phrases which contain keywords used in an unrelated context (e.g, ``in the spirit of [...]''). The combination of large, poorly accessible and noisy databases with search queries related to complex and domain-dependent topics such as R/S underscores the urgent need for systems that can quickly and easily retrieve, synthesize and generate knowledge based on complex and nuanced user queries, providing scholars and policymakers with low-barrier access to accurate and reliable information on complex topics. 

Current Large Language Models (LLMs) and general-purpose Information Retrieval (IR) systems often produce outputs that lack contextual sensitivity, thereby risking the homogenization of nuanced perspectives exemplified by the discourse on R/S \citep{hutchinson2024modeling, liu2024measuring, plaza2024divine}. Recent studies have highlighted that LLMs may exhibit contextual biases, particularly when predicting public opinion across diverse national and linguistic settings, leading to inaccuracies and a failure to capture cultural subtleties \citep{qu2024performance, von2024united}. Furthermore, evaluations of LLMs' cultural alignment reveal significant limitations, including instability across different cultural dimensions and challenges in consistently steering models to represent specific cultural perspectives \citep{masoud2025cultural, khan2025randomness}. These findings underscore the need for specialized knowledge systems capable of retrieving and generating knowledge in information environments marked by high context- and domain dependence and cultural complexity.
However, no existing systems were designed for knowledge acquisition in the context of R/S.

To address this gap, we present SpiritRAG—a Retrieval-Augmented Generation (RAG; \citet{lewis2020retrieval}) Question Answering (Q\&A) system designed to retrieve, synthesize, and generate R/S-related knowledge based on the resolutions stored in the UN archive. For present purposes, the corpus was built using data on health and education, with the latter included because many education-related documents are implicitly related to health (e.g., on tobacco control or sexual and reproductive care, see Sustainable Development Goal 3 \citep{UNSDG2024}). SpiritRAG is built on the top of approx.~7,500 documents available in seven languages, and supports interactive querying with real-time answers linked to exact reference documents. To evaluate the system, we compiled an expert-written test set of R/S-related questions in health and education. Pilot tests show that SpiritRAG is perceived by social scientists as a highly innovative and potentially groundbreaking interactive knowledge engine, highlighting its potential as a proof-of-concept for future LLM-augmented work in public policy, historical scholarship, and more generally, in the field of digital humanities.

Our main contributions are: (1) We developed an interactive Q\&A system for retrieving R/S-related knowledge; (2) We built a corpus of nearly 7,500 UN resolution documents on health and education subjects; (3) We created a test set of 100 evaluation questions authored by domain experts.

A video demonstration of SpiritRAG can be accessed at \url{https://youtu.be/x_G8bopwp8A?si=irrrlslw_i_ryRCX}.

\section{SpiritRAG: A Q\&A System for R/S}

SpiritRAG adopts a similar modular design principle as proposed by \citet{gao2024modoc} (see Figure~\ref{fig:pipeline}). In the following, we discuss the core components of the proposed system.

\subsection{Corpus}

\begin{figure}[htb]
    \centering
    \includegraphics[width=\linewidth]{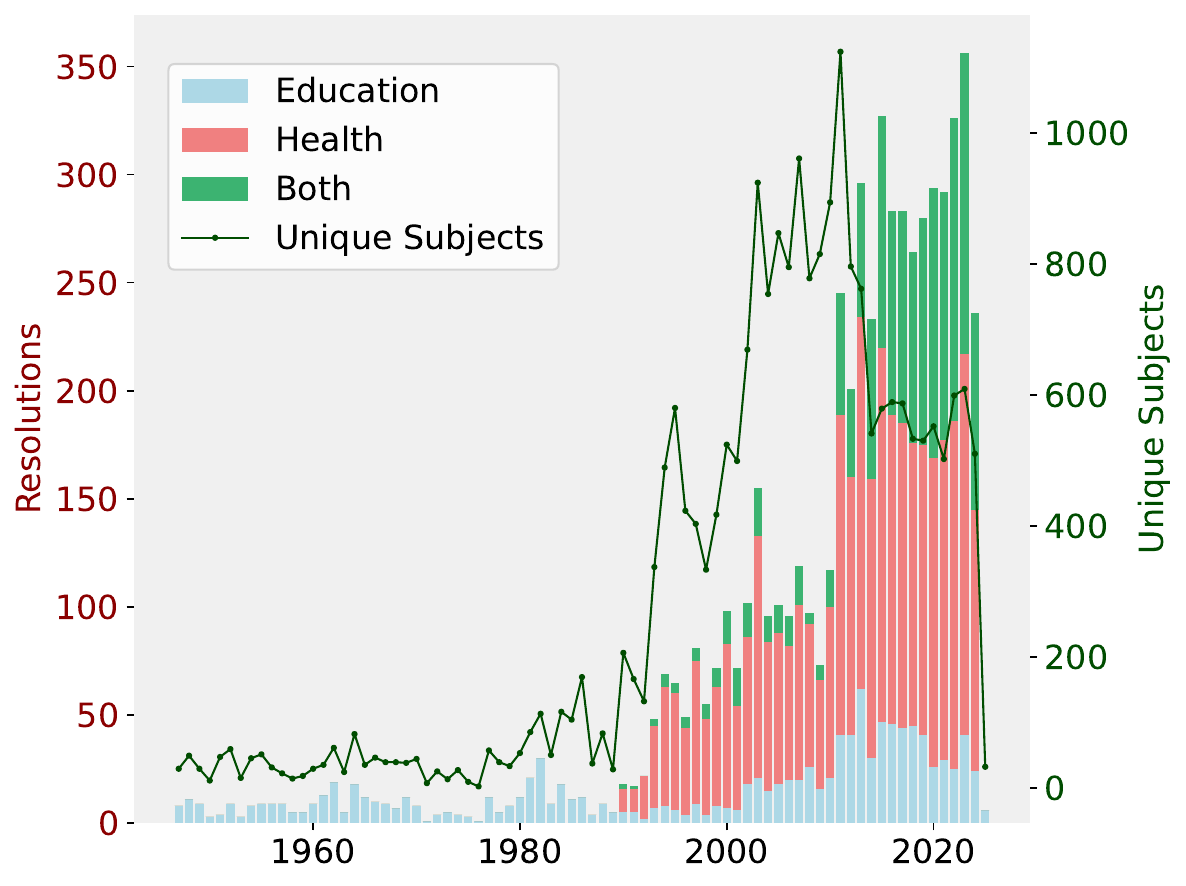}
    \caption{Resolutions and subjects across domains and decades in the UN-RES dataset.}
    \label{fig:subjects-overview}
\end{figure}

We constructed the UN-RES corpus by crawling the United Nations Official Document System\textsuperscript{\href{https://documents.un.org/}{2}} using a Python script with the web browsing simulator \texttt{Selenium}\footnote{\url{https://github.com/seleniumhq/selenium}, Apache-2.0 license.}. Specifically, we searched for UN documents (PDF files) that are: (1) resolution documents (marked RES); (2) related to the subjects \textit{health} and \textit{education}; and (3) published between \textit{01.01.1990} and \textit{03.31.2025}. This process resulted in a total of nearly 7,500 UN resolution documents across seven languages: \textit{Arabic} (ar), \textit{Chinese} (zh), \textit{English} (en), \textit{French} (fr), \textit{German} (de), \textit{Russian} (ru), and \textit{Spanish} (es). The overall dataset statistics are presented in Table~\ref{tab:stat}. Figure~\ref{fig:subjects-overview} visualizes the temporal distribution of UN resolution documents within the health and education domains, along with the corresponding counts of unique subjects.

\definecolor{spiritgreen}{HTML}{1C8D96}
\colorlet{lightgreen}{spiritgreen!60}  

\begin{table}[!htb]
    \centering
    \resizebox{\columnwidth}{!}{
    \begin{tabular}{lccccccc}
    \hline
    \textbf{Freq.} & \textbf{ar} & \textbf{de} & \textbf{en} & \textbf{es} & \textbf{fr} & \textbf{ru} & \textbf{zh} \\
    \hline
    \rowcolor{gray!20!white}
    \multicolumn{8}{c}{health + R/S} \\
    \# Doc. & 4,669 & 2,328 & 4,781, & 4,670 & 4,747 & 4,690 & 4,759 \\
    \# Subj. & 2,360 & 2,046 & 2,383 & 2,361 & 2,377 & 2,367 & 2,380 \\
    \hline
    \rowcolor{gray!20!white}
    \multicolumn{8}{c}{education} \\
    \# Doc. & 2,586 & 1,123 & 2,718 & 2,658 & 2,694 & 2,670 & 2,586 \\
    \# Subj. & 1,865 & 1,455 & 1,944 & 1,919 & 1,935 & 1,930 & 1,938 \\
    \hline
    \end{tabular}
    }
    \caption{Frequency of documents and subjects per language for each subject domain.}
    \label{tab:stat}
\end{table}

To provide an initial overview of the data, we conducted a hierarchical cluster analysis of the 2,652 unique subject tags included in the metadata of the UN-RES corpus. The resulting 185 concise and interpretable clusters allowed for deeper insights into the temporal distribution of religious themes within the contexts of health and education. Figure~\ref{fig:education-clusters} in Appendix~\ref{sec:subjects} shows the average weight fluctuations of selected relevant clusters across decades.~Figure~\ref{fig:religion-cluster} presents a more fine-grained analysis of specific religion-related subjects, broken down by five-year periods.

We further parsed crawled documents in English, French, German, and Spanish into a structured JSON format with paragraph chunks using the document parsing toolkit \texttt{Docling}\footnote{\url{https://github.com/docling-project/docling}, MIT license.} \citep{Docling}\footnote{We did not parse other languages due to the lack of high-quality PDF parsers.}. From the parsed documents, we constructed the vector database for UN-RES with the efficient nearest neighbor search toolkit \texttt{Faiss}\footnote{\url{https://github.com/facebookresearch/faiss}, MIT license.}, by using the pre-trained text encoder Sentence-BERT \citep{reimers2019sentence} and Qwen3-Embedding \citep{zhang2025qwen3embedding} to pre-compute the document embeddings.

\subsection{Models}

\paragraph{Document retrieval.} Given a user query $q$ (i.e., a R/S-related question), we use the same text encoder to compute the query embedding $\mathbf{e}_q$ on the fly. The document retriever first computes the cosine similarity score between the query embedding $\mathbf{e}_q$ to each pre-computed document embedding $\mathbf{e}_d \in \mathbf{E}^{|D| \times m}$ (where $m$ denotes the embedding dimensionality, $D$ the corpus, and $d$ the document in $D$) stored in the database: 
\begin{align*}
    \text{cos\_sim}(\mathbf{e}_q, \mathbf{e}_d) = \frac{\mathbf{e}_q \cdot \mathbf{e}_d}{||\mathbf{e}_q|| \cdot ||\mathbf{e}_d||}.
\end{align*}
Based on the resulting similarity scores, the retriever pre-fetches the top-$n$ most relevant resolution documents $(D_i)_{i=1}^n$ by ranking them in descending order of cosine similarity (i.e., the higher the similarity, the higher the rank):
\begin{equation}
\resizebox{\columnwidth}{!}{%
\begin{minipage}{\columnwidth}
\vspace{-1.5em}
\begin{align*}
& \text{rank}(d) = \\
& |\{ d' \in D\ |\ \text{cos\_sim}(\mathbf{e}_q, \mathbf{e}_{d'}) 
> \text{cos\_sim}(\mathbf{e}_q, \mathbf{e}_d) \}| + 1.
\end{align*}
\end{minipage}%
}
\end{equation}
The document retriever then re-ranks the documents in $(D_i)_{i=1}^n$ by calculating the similarity between the query embedding and each sentence embedding $\mathbf{e}_s$ within a given $D_i$, using both the maximal and average similarities. Each $D_i$ is re-ranked according to the relevance score
\begin{align*}
        & r(D_i;q) = \\
        & \alpha \cdot \max_{s \in D_i}(L_2(\mathbf{e}_q, \mathbf{e}_s))
        + (1-\alpha) \cdot \operatorname*{avg}_{s \in D_i} (L_2(\mathbf{e}_q, \mathbf{e}_s)),
\end{align*}
where $s$ is a sentence in $D_i$, $L_2$ denotes the Euclidean distance and $\alpha$ is a weighting factor that was empirically set to 0.7. The final retrieved documents are the top-$k$ entries in the re-ranked list.

\begin{figure}[htb]
\centering
\begin{tcolorbox}[width=\columnwidth,  
  colback=gray!5!white,
  colframe=gray!75!black,
  fontupper=\normalsize
]
You are a helpful AI assistant. Use the following information to answer the user's question. \\

User's question: \{query\} \\

Relevant information from the retrieved documents: \{retrieved\_docs\} \\

Relevant information from the user uploaded PDF (optional): \{parsed\_pdf\}
\end{tcolorbox}
\caption{Prompt template used for answer generation.}
\label{fig:ans-prompt}
\end{figure}

\paragraph{Answer generation.} Conditioned on the user query and the top-$k$ most relevant documents retrieved from the database, we use instruction-tuned Qwen3 models \citep{yang2025qwen3} to generate the answer by assembling the inputs with a simple prompt template (see Figure~\ref{fig:ans-prompt}) and passing them to the model.

In practice, users often have reliable source documents that may provide additional context for their questions. To support this, we implemented a function that allows users to upload a source PDF file, which is parsed and chunked in real time. When available, this source file is also included in the conditional generation of the answer.

To accelerate the generation process, we implemented the answer generation with the efficient inference toolkit \texttt{vLLM}\footnote{\url{https://github.com/vllm-project/vllm}, Apache-2.0 license.} \citep{kwon2023efficient}.

\subsection{Demonstration}

SpiritRAG has a front-end implemented with \texttt{React.js}\footnote{\url{https://react.dev/}} and a back-end engineered using \texttt{Flask}\footnote{\url{https://flask.palletsprojects.com/}} APIs. With this modular design, SpiritRAG is lightweight and easily deployed both locally or on cloud services, requiring minimal computation resources. Our pilot system was deployed on a university-internal cloud service using an NVIDIA Tesla L4 GPU card with 24 GB of RAM.

\begin{figure*}[htb]
    \centering
    \includegraphics[width=\linewidth]{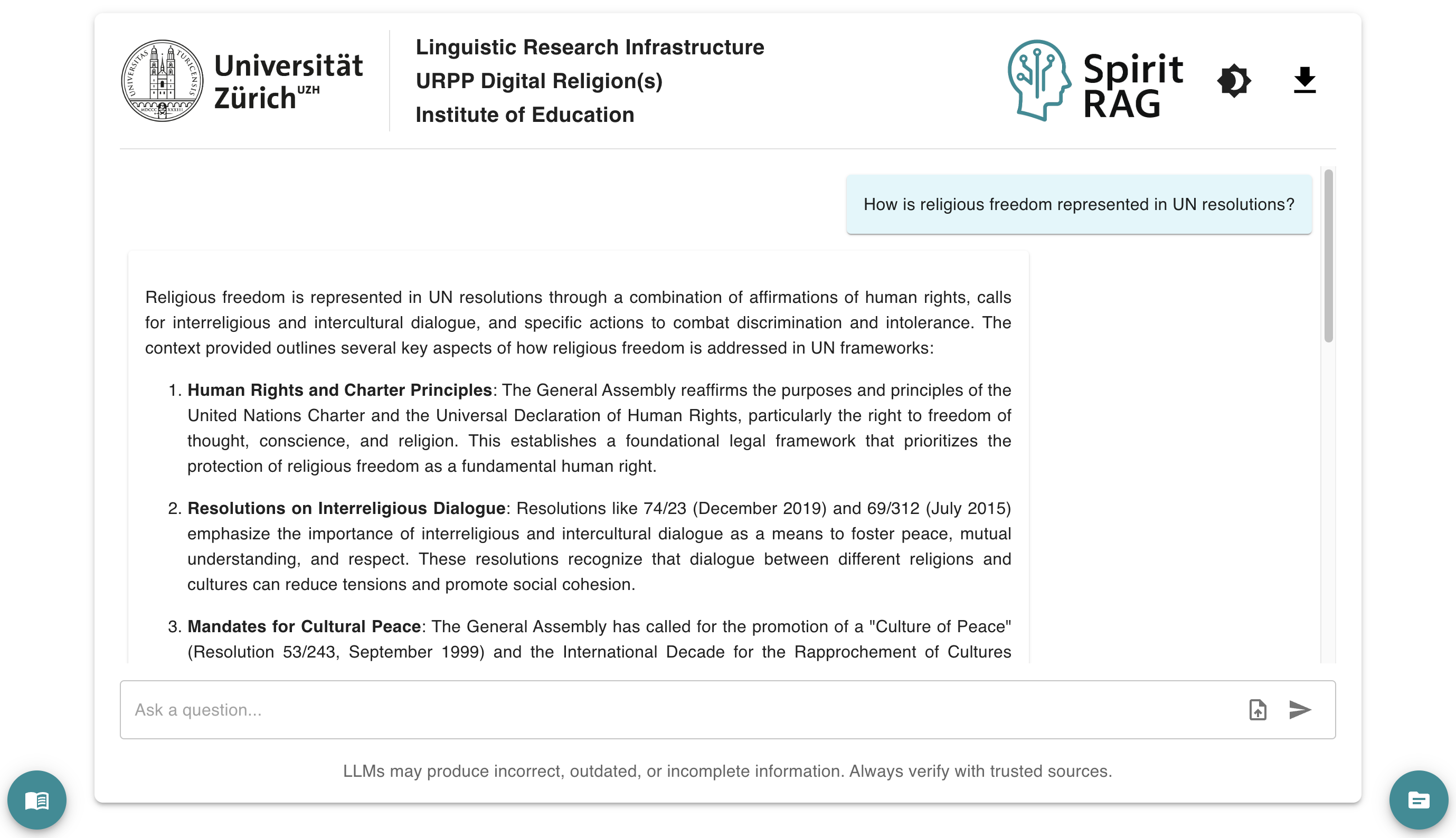}
    \caption{Overview of SpiritRAG. Generated responses are rendered in Markdown to improve readability of long answers. This example answer was generated by \texttt{Qwen3-1.7B} with \texttt{Qwen3-Embedding-0.6B} as document retriever.}
    \label{fig:ui-overall}
\end{figure*}

Figure~\ref{fig:ui-overall} demonstrates the user interface of SpiritRAG. SpiritRAG allows real-time human-machine interactions in a chatbot setting, where users can freely ask questions related to R/S. The answers generated by the Qwen3 model are rendered in Markdown for improved readability. Users can switch to dark mode and download the conversation history by clicking on the buttons in the banner.

On the right panel of the web application, accessible via the source button in the lower right corner, the most relevant documents used for answer generation are displayed (see Figure~\ref{fig:right-panel} in Appendix~\ref{sec:interface}). Each retrieved document is presented with its metadata, including document title, key subjects (highlighted in \textcolor{spiritgreen}{green}), publication date, and available languages. We report more statistics on the subjects in Appendix~\ref{sec:subjects}.

When one of the available languages is clicked, a PDF viewer will be opened in the left panel, displaying the original UN resolution document (see Figure~\ref{fig:left-panel} in Appendix~\ref{sec:interface}). The left panel can be closed by clicking on any empty area on the web application.

\begin{table*}[!htb]
    \centering
    \resizebox{0.9\textwidth}{!}{
    \begin{tabular}{lccccc}
    \hline
     \textbf{Document Retriever} & \textbf{Relevance} & \textbf{Accuracy} & \textbf{Usefulness} & \textbf{Temporality} & \textbf{Actionability}  \\
     \hline
     \texttt{all-MiniLM-L6-v2} & 3.78 & 4.45 & 3.66 & 4.45 & \textcolor{spiritgreen}{2.57}  \\
     \rowcolor{gray!20!white}
     \texttt{Qwen3-Embedding-0.6B} & \textcolor{spiritgreen}{4.26} & \textcolor{spiritgreen}{4.99} & \textcolor{spiritgreen}{3.89} & \textcolor{spiritgreen}{4.99} & 2.08 \\
     \hline
     \textbf{Answer Generator} & \textbf{Congruence} & \textbf{Coherence} & \textbf{Relevance} & \textbf{Creativity} & \textbf{Engagement}  \\
     \hline
    \texttt{Qwen3-0.6B} & 2.27 & 3.29 & 2.44 & 1.94 & 2.06 \\
    \rowcolor{gray!20!white}
    \texttt{Qwen3-1.7B} & \textcolor{spiritgreen}{3.41} & \textcolor{spiritgreen}{3.94} & \textcolor{spiritgreen}{4.18} &\textcolor{spiritgreen}{3.39} & \textcolor{spiritgreen}{2.51} \\
    \hline
    \end{tabular}
    }
    
    \caption{Human evaluation results on health + R/S related test questions.}
    \label{tab:evaluation-health}
\end{table*}

\begin{table*}[!htb]
    \centering
    \resizebox{0.9\textwidth}{!}{
    \begin{tabular}{lccccc}
    \hline
     \textbf{Document Retriever} & \textbf{Relevance} & \textbf{Accuracy} & \textbf{Usefulness} & \textbf{Temporality} & \textbf{Actionability}  \\
     \hline
     \texttt{all-MiniLM-L6-v2} & 2.56 & 2.57 & 2.44 & 4.70 & 2.38  \\
     \rowcolor{gray!20!white}
     \texttt{Qwen3-Embedding-0.6B} & \textcolor{spiritgreen}{3.51} & \textcolor{spiritgreen}{3.43} & \textcolor{spiritgreen}{3.42} & \textcolor{spiritgreen}{4.75} & \textcolor{spiritgreen}{3.39} \\
     \hline
     \textbf{Answer Generator} & \textbf{Congruence} & \textbf{Coherence} & \textbf{Relevance} & \textbf{Creativity} & \textbf{Engagement}  \\
     \hline
     \texttt{Qwen3-0.6B} & 2.42 & 2.71 & 2.23 & 3.54 & 2.71  \\
    \rowcolor{gray!20!white}
    \texttt{Qwen3-1.7B} & \textcolor{spiritgreen}{3.78} & \textcolor{spiritgreen}{3.76} & \textcolor{spiritgreen}{3.98} & \textcolor{spiritgreen}{3.92} & \textcolor{spiritgreen}{4.06} \\
    \hline
    \end{tabular}
    }
    
    \caption{Human evaluation results on education related test questions.}
    \label{tab:evaluation-education}
\end{table*}

\section{Evaluation}


We performed human evaluation of SpiritRAG's document retrieval and answer generation capabilities. For this purpose, we asked two domain experts specialized in R/S and education research to assess SpiritRAG along multiple evaluation dimensions. 

For each question in the test of 100 questions (see Appendix~\ref{sec:questions}), the domain experts were first asked to evaluate the retrieved documents. After an answer was generated, experts could activate the assessment mode by clicking the Eval button located at the top right of the right panel. For each retrieved document and the generated answer, a five-point Likert scale was provided to assess six evaluation dimensions (see Appendix~\ref{sec:questions} for details).

\section{Results and Discussion}

In Table~\ref{tab:evaluation-health} and Table~\ref{tab:evaluation-education}, we report the average ratings across each evaluation dimension. We observe that the quality of both retrieved documents and generated answers improves as the backbone document retriever and answer generator are replaced with larger language models. Evaluators noted the considerable improvement in performance between Sentence-BERT and Qwen3-Embedding as document retriever, and particularly a significant and consistent improvement in answer length and detail of \texttt{Qwen3-0.6B} in comparison with \texttt{Qwen3-1.7B} as answer generator. It is worth noting that, in the case of document retrieval, a low actionability score does not necessarily indicate poor retriever performance. Rather, it reflects the fact that only a limited number of UN resolutions contain actionable directives, and ambivalence regarding how actionability should be understood in the context of several evaluation questions. 

A more detailed, iterative strategy of evaluation and fine-tuning based on systematically developed evaluation dimensions tailored to the corpus was suggested to further strengthen the feasibility of SpiritRAG as  a widely used research tool. An informal pilot in an interdisciplinary academic setting suggested that SpiritRAG is perceived by subject experts as a highly innovative, interdisciplinary proof-of-concept with the potential for far-reaching impact in their respective academic disciplines.

\noindent
\section{Related Works}
\label{sec:related-works}
SpiritRAG is relevant to scholars in religious studies, health policy, and international relations. It is of particular interest to researchers interested in methodological innovation in the study of religion \citep{wright_methodological_2022}, the use of digital humanities methods in the study of religion \citep{hutchings_digital_2015} and the use of neural networks for text analysis \citep{kim2020dimensions, suissa_text_2022}. SpiritRAG also provides a workable proof-of-concept for theologians interested in exploring new forms of dialogue-based interaction with religious texts \citep{Kurlberg2020Doing, garner2021theology, sutinen2021digital}. SpiritRAG complements computational analyses of large text corpora in fields that have mostly focused on topic modeling to date \citep{Park2023UNESCO, Yamada2024synchronic} and proposes a new resource for scholars looking at inter-textual discursive relationships in global governance and social policy \citep{Windzio2022Talking, Montjourides2023Is}.

\section{Conclusion and Future Work}

We presented SpiritRAG, a Q\&A system enhanced with RAG for the domains of health plus R/S within the context of United Nations resolution documents. Built on top of approx.~7,500 UN resolution documents, SpiritRAG is designed to support social scientists and policymakers in the initial stages of knowledge acquisition by enabling interactive retrieval of contextualized, domain-dependent information. This, in turn, eases evidence-informed research and decision-making. In future work, we plan to expand SpiritRAG to incorporate a broader range of document sources, including religious texts, with the goal of making the embedded spiritual knowledge in these documents more accessible and relevant to contemporary discourse.

\section*{Limitations}

SpiritRAG has several limitations. First, due to computational constraints, we did not evaluate larger-scale LLMs for document retrieval and answer generation. Second, in the absence of ground-truth answers and relevant documents, we did not fine-tune the document retriever for optimal performance. Third, SpiritRAG is currently tailored to address education- or health + R/S-related questions within the context of UN resolution documents only; it does not incorporate documentation from other UN organizations and related multilateral organizations. Nevertheless, thanks to its modular architecture, SpiritRAG can be readily extended to other domain-specific knowledge sources.

\section*{Acknowledgments}
This work is funded by the URPP Digital Religion(s) at the University of Zurich (UZH) and the Swiss National Science Foundation, grant number 204286, with extensive support provided by the Chair of History of Education and Education Policy Analysis at  Institute of Education of UZH. We also acknowledge NCCR Evolving
Language, Swiss National Science Foundation Agreement No. 51NF40\_180888. A special thanks to Mr. Igor Mustač at LiRI for deploying SpiritRAG online.

\bibliography{custom}

\appendix
\onecolumn


\section{Test Set Questions and Evaluation Dimensions}
\label{sec:questions}

\begin{table}[H]
    \centering
    \resizebox{\textwidth}{!}{
    \begin{tabular}{cp{11.5cm}cc}
    \hline
    \textbf{ID} & \textbf{Question}  & \textbf{Type} &  \textbf{TB?} \\
    \hline
    \multirow{2}{*}{1} & How has the relationship between religion, spirituality, and health been framed in policy texts adopted since 1946? & \multirow{2}{*}{Open-ended} & \multirow{2}{*}{Yes} \\
    \multirow{2}{*}{2} & What references are made to gender equality in relation to religion, spirituality across UN resolutions? & \multirow{2}{*}{Open-ended} & \multirow{2}{*}{No} \\
    3 & Which resolutions cite religion and/or spirituality? & List (finite) & No \\
    4 & Have any UN resolutions addressed the right to religious freedom? & Yes/No & No \\
    5 & What provisions have been made for refugees' religious freedom? & Open-ended* & No \\
    6 & How is the phrase ``culturally appropriate'' used across UN documents? & Open-ended* & No \\
    7 & Do UN documents address the notion of human dignity? & Yes/No & No \\
    8 & What emphasis is placed on the notion of human dignity? & Open-ended & No \\
    \multirow{2}{*}{9} & When was United Nations Interagency Task Force on Religion and Sustainable Development incorporated in UN resolutions? & \multirow{2}{*}{Closed} & \multirow{2}{*}{Yes} \\
    \multirow{2}{*}{10} & What UN Member States commitments have been made toward religious freedom? & \multirow{2}{*}{Open-ended*} & \multirow{2}{*}{No} \\
    11 & How often is religious freedom mentioned in UN resolutions since 1990? & Closed & No \\
    \multirow{2}{*}{12} & How has UN language associated with religion and/or spirituality evolved over time? & \multirow{2}{*}{Open-ended*} & \multirow{2}{*}{Yes} \\
    \multirow{2}{*}{13} & What international monitoring frameworks are proposed for religious freedom-related targets? & \multirow{2}{*}{List (finite)} & \multirow{2}{*}{Yes} \\
    14 & How is religious freedom represented in UN resolutions? & Open-ended & Yes \\
    \multirow{2}{*}{15} & Have there been calls for interreligious or interfaith dialogue in UN texts and resolution? & \multirow{2}{*}{Yes/No} & \multirow{2}{*}{No} \\ 
    16 & Are there any specific resolutions that discuss religion or spirituality? & List (finite) & No \\
    \multirow{2}{*}{17} & How does the UN address religion and/or spirituality in conflict-affected regions? & \multirow{2}{*}{Open-ended} & \multirow{2}{*}{Yes} \\
    18 & Which resolutions link religion and/or spirituality to UN activities? & List (finite) & No \\
    19 & Are there references to faith-based organizations in UN resolutions? & Yes/No & No \\
    20 & How have UN texts and resolutions evolved in cultural appropriateness? & Open-ended* & Yes \\
    \multirow{2}{*}{21} & How have religious leaders, communities or actors contributed to the Sustainable Development Goals? & \multirow{2}{*}{Open-ended*} & \multirow{2}{*}{No} \\
    22 & Does any UN text or resolution cite cultural appropriateness as a priority? & Yes/No & No \\
    \multirow{2}{*}{23} & Is there any reference to religious involvement in UN-related activities in recent UN texts and resolutions? & \multirow{2}{*}{Yes/No} & \multirow{2}{*}{Yes} \\
    24 & Is interfaith or interreligious dialogue addressed by the United Nations? & Yes/No & No \\
    25 & What resolutions discuss interfaith or interreligious education? & List (finite) & No \\
    26 & Have any resolutions referred to culturally appropriate policies? & Yes/No & No \\
    27 & Is health framed as a human right consistently? & Open-ended & No \\
    \multirow{2}{*}{28} & What are the main dimensions of religious discrimination acknowledged in UN resolutions? & \multirow{2}{*}{List (finite)} & \multirow{2}{*}{Yes} \\
    \multirow{2}{*}{29} & Does any UN organizations take a specific interest in religion and/or spirituality? & \multirow{2}{*}{Yes/No} & \multirow{2}{*}{No} \\
    \multirow{2}{*}{30} & How well do UN texts and resolutions address issues pertaining to public expenditure on healthcare? & \multirow{2}{*}{Open-ended*} & \multirow{2}{*}{No} \\
    \multirow{2}{*}{31} & How do UN resolution approach healthcare for persons in low- or middle income countries? & \multirow{2}{*}{Open-ended*} & \multirow{2}{*}{No} \\
    \multirow{2}{*}{32} & What frameworks have been proposed over time to monitor equity in healthcare? & \multirow{2}{*}{List (finite)} & \multirow{2}{*}{No} \\
    \hline
    \end{tabular}
    }
    \caption{Health + R/S test questions (part 1).}
    \label{tab:health-question-1}
\end{table}

\begin{table}[H]
    \centering
    \resizebox{\textwidth}{!}{
    \begin{tabular}{cp{11.5cm}cc}
    \hline
    \textbf{ID} & \textbf{Question}  & \textbf{Type} &  \textbf{TB?} \\
    \hline
    \multirow{2}{*}{33} & How often has ``universal primary healthcare'' as a global or priority been emphasized in the past decade? & \multirow{2}{*}{Open-ended} & \multirow{2}{*}{Yes} \\
    34 & Is ``Health for All'' referenced in UN resolutions? & Yes/No & No \\
    \multirow{2}{*}{35} & What kind of healthcare system reforms are promoted by United Nations organizations? & \multirow{2}{*}{Open-ended*} & \multirow{2}{*}{No} \\
    36 & What role is assigned to UN agencies like WHO in healthcare? & Open-ended* & No  \\
    37 & How is health in emergencies addressed by the United Nations? & Open-ended & Yes \\
    \multirow{2}{*}{38} & Do any resolutions define global priorities for healthcare? And if so, how are they defined? & \multirow{2}{*}{Yes/No + Open-ended} & \multirow{2}{*}{No} \\
    39 & What provisions exist for global health emergencies in UN official texts? & Open-ended & No \\
    40 & Are health disparities based on socioeconomic status addressed by the UN? & Yes/No & No \\
    \multirow{2}{*}{41} & How have the United Nations addressed racial or ethnic discrimination in healthcare over time? & \multirow{2}{*}{Open-ended*} & \multirow{2}{*}{No} \\
    42 & Have the Untied Nations paid attention to healthcare in rural areas? & Yes/No & No \\
    43 & To which extent do UN resolutions support healthcare education? & Open-ended & No \\
    \multirow{2}{*}{44} & How have the UN embedded healthcare in their mandate to foster peaceful societies? & \multirow{2}{*}{Open-ended} & \multirow{2}{*}{No} \\
    45 & What resolutions address data collection in healthcare systems? & List (finite) & No \\
    46 & Are there mentions of global pandemics in health? & Yes/No & No \\
    47 & How do resolutions define the right to health? & Open-ended & Yes \\
    \multirow{2}{*}{48} & What role does healthcare play in sustainable development according to the UN? & \multirow{2}{*}{Open-ended} & \multirow{2}{*}{Yes} \\
    \multirow{2}{*}{49} & Are gender-based discrimination in healthcare among the concerns voiced by UN Member States? & \multirow{2}{*}{Yes/No} & \multirow{2}{*}{No} \\ 
    50 & What resolutions mention health as a human right? & List (finite) & No \\
    \hline
    \end{tabular}
    }
    \caption{Health + R/S test questions (part 2).}
    \label{tab:health-question-2}
\end{table}

\vspace{-0.5em}
\begin{table}[H]
    \centering
    \resizebox{\textwidth}{!}{
    \begin{tabular}{cp{11.5cm}cc}
    \hline
    \textbf{ID} & \textbf{Question}  & \textbf{Type} &  \textbf{TB?} \\
    \hline
    1 & How has the right to education been framed in UN texts adopted since 1946?  & Open-ended & Yes \\
    \multirow{2}{*}{2} & What references are made to gender equality in education across UN resolutions? & \multirow{2}{*}{Open-ended} & \multirow{2}{*}{No} \\
    3 & Which resolutions discuss access to primary education? & List (finite) & No \\
    4 & Have any UN resolutions addressed the digital divide in education? & Yes/No & No \\
    5 & What provisions have been made for refugees' education? & Open-ended* & No \\
    6 & How is inclusive education defined in UN resolutions? & Open-ended* & No \\
    7 & Do UN resolutions address special needs education? & Yes/No & No \\
    \multirow{2}{*}{8} & What emphasis is placed on technical and vocational education in UN documents between 1960 and 1970? & \multirow{2}{*}{Open-ended} & \multirow{2}{*}{Yes} \\
    9 & When was lifelong learning incorporated in UN resolutions? & Closed & No \\
    \multirow{2}{*}{10} & What UN Member States commitments have been made toward teacher training and professional development? & \multirow{2}{*}{Open-ended*} & \multirow{2}{*}{No} \\
    \multirow{2}{*}{11} & How often is early childhood education mentioned in UN resolutions since 1981? & \multirow{2}{*}{Closed} & \multirow{2}{*}{Yes} \\
    \multirow{2}{*}{12} & How has the UN language associated with justice and fairness in education evolved over time? & \multirow{2}{*}{Open-ended*} & \multirow{2}{*}{No} \\
    \multirow{2}{*}{13} & What international monitoring frameworks are proposed for education-related targets? & \multirow{2}{*}{List (finite)} & \multirow{2}{*}{Yes} \\
    14 & How are indigenous peoples’ education rights represented in UN resolutions? & Open-ended & Yes \\
    15 & Have there been calls for universal free education in UN texts and resolution? & Yes/No & No \\
    16 & Are there any specific resolutions that define quality education? & List (finite) & No \\
    17 & How does the UN address education in conflict-affected regions? & Open-ended & Yes \\
    \hline
    \end{tabular}
    }
    \caption{Education-related test questions (part 1).}
    \label{tab:edu-question-1}
\end{table}

\begin{table}[H]
    \centering
    \vfill
    \resizebox{\textwidth}{!}{
    \begin{tabular}{cp{11.5cm}cc}
    \hline
    \textbf{ID} & \textbf{Question}  & \textbf{Type} &  \textbf{TB?} \\
    \hline
    18 & Which resolutions link education and peace building? & List (finite) & No \\
    19 & Are there references to education financing in UN resolutions? & Yes/No & No \\
    20 & How have UN texts and resolutions evolved in support for girls’ education? & Open-ended* & No \\
    21 & How has the SDG 4 agenda been further reflected in resolutions since 2015? &  Open-ended* & Yes \\
    22 & Does any UN text or resolution cite learning outcomes as a priority? & Yes/No & No \\
    \multirow{2}{*}{23} & Is there any reference to private sector involvement in education in recent UN texts and resolutions? & \multirow{2}{*}{Yes/No} & \multirow{2}{*}{Yes} \\
    24 & Are online and distance learning modalities addressed by the United Nations? & Yes/No & No \\
    25 & What resolutions discuss multilingual education? & List (finite) & No \\
    26 & Have any resolutions referred to climate change education? & Yes/No & No \\
    27 & Is education framed as a human right consistently? & Open-ended & No \\
    \multirow{2}{*}{28} & What are the main dimensions of exclusion acknowledged in UN resolutions during the Cold War? & \multirow{2}{*}{List (finite)} & \multirow{2}{*}{Yes} \\
    29 & Does UNESCO recommend specific pupils-teacher ratios? & Yes/No & No \\
    \multirow{2}{*}{30} & How well do UN texts and resolutions address issues pertaining to public expenditure on education? & \multirow{2}{*}{Open-ended*} & \multirow{2}{*}{No} \\
    31 & How do UN resolutions approach education for persons with disabilities? & Open-ended* & No \\
    \multirow{2}{*}{32} & What are the frameworks proposed in UN resolution to monitor progress towards equity in education? & \multirow{2}{*}{List (finite)} & \multirow{2}{*}{No} \\
    \multirow{2}{*}{33} & How often has literacy as a goal or priority been emphasized in the past decade? & \multirow{2}{*}{Open-ended} & \multirow{2}{*}{Yes} \\
    34 & Is citizenship or civic education referenced in UN resolutions? & Yes/No & No \\
    35 & What kind of education system reforms are promoted by the United Nations? & Open-ended* & No \\
    36 & What role is assigned to UN agencies like UNESCO in education? & Open-ended* & No \\
    37 & How is education in emergencies addressed by the United Nations? & Open-ended & Yes \\
    \multirow{2}{*}{38} & Do any resolutions define global citizenship education? And if so, how is it defined? & \multirow{2}{*}{Yes/No + Open-ended} & \multirow{2}{*}{No} \\
    39 & What provisions exist for non-formal education in UN official texts? & Open-ended & No \\
    \multirow{2}{*}{40} & Are educational disparities based on socioeconomic status addressed by the UN? & \multirow{2}{*}{Yes/No} & \multirow{2}{*}{No} \\
    \multirow{2}{*}{41} & How have the United Nations addressed racial or ethnic discrimination in education over time? & \multirow{2}{*}{Open-ended*} & \multirow{2}{*}{No} \\
    42 & Have the United Nations paid attention to education in rural areas? & Yes/No & No \\
    43 & To which extent do UN resolutions support bilingual education? & Open-ended & No \\
    \multirow{2}{*}{44} & How have the UN embedded education in their mandate to foster peaceful societies? & \multirow{2}{*}{Open-ended} & \multirow{2}{*}{No} \\
    45 & What resolutions address data collection in education systems? & List (finite) & No \\
    46 & Are there mentions of international assessments in education? & Yes/No & No \\
    47 & How do resolutions define access to education? & Open-ended & Yes \\
    \multirow{2}{*}{48} & What role does education play in sustainable development according to the UN? & \multirow{2}{*}{Open-ended} & \multirow{2}{*}{Yes} \\
    \multirow{2}{*}{49} & Are gender stereotypes in curricula among the concerns voiced by UN Member States? & \multirow{2}{*}{Yes/No} & \multirow{2}{*}{No} \\
    50 & What resolutions mention teacher rights and working conditions? & List (finite) & No \\
    \hline
    \end{tabular}
    }
    \caption{Education-related test questions (part 2).}
    \label{tab:edu-question-2}
\end{table}

\vspace{-2em}
\begin{table}[H]
    \centering
    \begin{tabular}{lp{13cm}}
    \hline
     \textbf{Type} &  \textbf{Definition} \\
     \hline
     List (finite) & Desired answers are in form of finite list with bullet points. \\
     Yes/No & Questions that can be answered in binary form. \\
     Closed & Questions that can be precisely answered. \\
     Open-ended & No special format requirement for desired answers. \\
     Open-ended* & No special format requirement for desired answers, yet diachronic changes possible. \\
     
     \hline
    \end{tabular}
    \caption{Definition of question types.}
    \label{tab:type}
\end{table}
\begin{table}[H]
    \centering
    \resizebox{\linewidth}{!}{
    \small
    \begin{tabular}{ll}
    \hline
    \textbf{Dimension} & \textbf{Descriptions} \\
    \hline
    \rowcolor{gray!20!white}
    \multicolumn{2}{c}{Document Retrieval} \\
    Relevance & How relevant is the retrieved document to the question? \\
    Accuracy & How trustworthy is the retrieved document for answering the question?  \\
    Usefulness & How useful is the retrieved document for answering the question? \\
    Temporality & Is the retrieved document related to the time period implied by the question? \\
    Actionability & How many actionable insights does the retrieved document contain? \\
    \hline
    \rowcolor{gray!20!white}
    \multicolumn{2}{c}{Answer Generation} \\
    Congruence & How well does the answer align with the retrieved documents? \\
    Coherence & How logically consistent is the answer? \\
    Relevance & How relevant is the answer to the question? \\
    Creativity & How original and creative is the answer? \\
    Engagement & How engaging and interesting is the answer? \\
    \hline
    \end{tabular}
    }
    
    \caption{Evaluation dimensions for the retrieved documents and generated answer.}
    \label{tab:dimensions}
\end{table}

\section{Dataset Construction Details}
\label{sec:details}

\begin{figure}[H]
    \centering
    \includegraphics[width=\columnwidth]{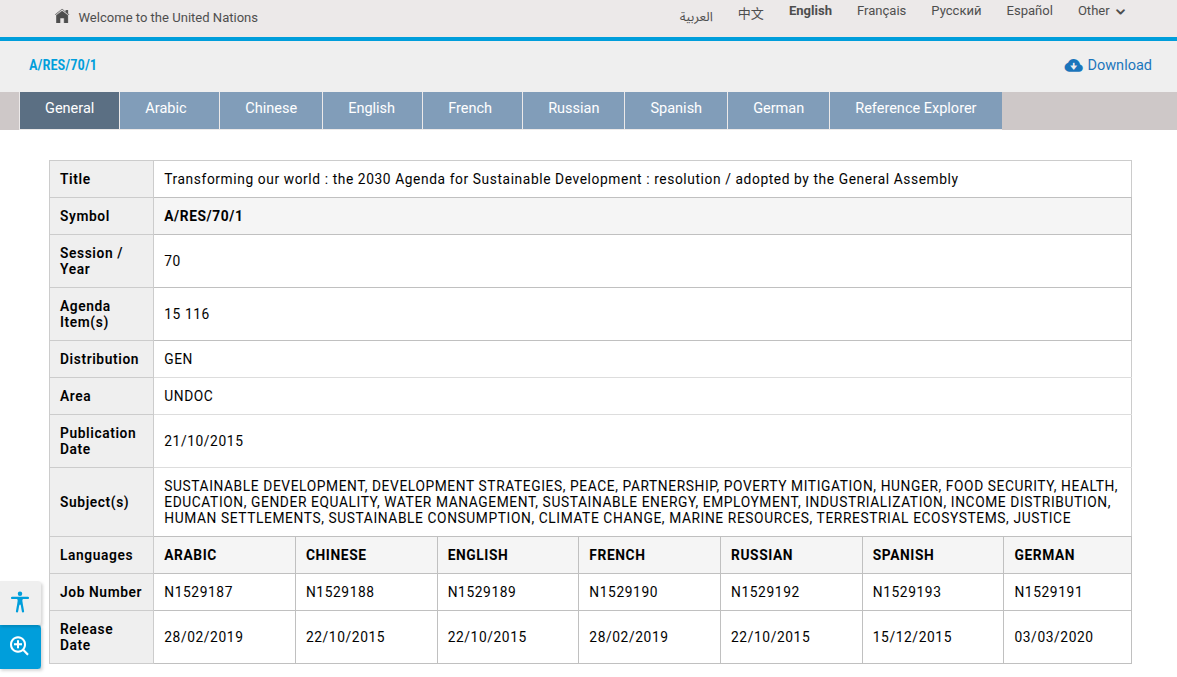}
    \caption{An example resolution document that is both health- and education-related. Screenshot taken from the website of UN Official Document System.}
    \label{fig:intersect}
\end{figure}

We implemented the web crawler using \texttt{Selenium} in combination with \texttt{BeautifulSoup}\footnote{\url{https://www.crummy.com/software/BeautifulSoup/bs4/doc/}, MIT License.}. Boolean operators were employed to combine keyword sets for document retrieval: \textit{educat*}, \textit{school*}, and \textit{learn*} for education-related documents, and \textit{health}, \textit{faith}, \textit{religi*}, \textit{spiritual*}, and \textit{belief} for documents related to health and religion/spirituality (R/S). Notably, some resolution documents are relevant to both domains; see Figure~\ref{fig:intersect} for an illustrative example.

\vspace{1em}
\section{Examples of Generated Answers}
\label{sec:example}

\begin{table}[H]
\centering
\resizebox{\textwidth}{!}{
\begin{tabular}{p{\columnwidth}}
\hline
\textbf{Q}: What references are made to gender equality in relation to religion, spirituality across UN resolutions? \\
\hline
 \rowcolor{gray!20!white} \multicolumn{1}{c}{\textcolor{qwen-embedding}{\texttt{Qwen3-Embedding-0.6B}} for document retrieval and \textcolor{qwen3-small2}{\texttt{Qwen3-1.7B}} for answer generation} \\
\textbf{A}: The references to gender equality in relation to religion, spirituality, and the promotion of interreligious and intercultural dialogue are embedded within several UN resolutions, particularly those addressing cultural pluralism, human rights, and the prevention of violence and discrimination. Here is a detailed breakdown of the key references: \newline

1. \textbf{Promotion of Interreligious and Intercultural Dialogue}

\quad \textbf{Resolution 75/26} (adopted on 2 December 2020) emphasizes the importance of interreligious and intercultural dialogue for peace, mutual understanding, and social stability. It underscores the need to create an environment conducive to peace and mutual respect at both global and local levels. This resolution directly ties gender equality to religious and spiritual diversity, highlighting the role of dialogue in fostering inclusive societies.\newline

2. \textbf{Human Rights and Religious Minorities} 

\quad \textbf{Resolution 74/306} (adopted on 11 September 2020) recognizes the need to counter discrimination, hate speech, and xenophobia related to the COVID-19 pandemic. It reaffirms the primary responsibility of states to protect human rights, including the rights of religious minorities. This resolution explicitly links gender equality to the protection of religious minorities, ensuring that women and other marginalized groups are not disproportionately affected by discrimination based on religion or belief. \newline

3. \textbf{Declaration on the Elimination of All Forms of Intolerance and Discrimination Based on Religion or Belief}

\quad \textbf{Resolution 36/55} (adopted on 25 November 1981) proclaims the \textbf{Declaration on the Elimination of All Forms of Intolerance and Discrimination Based on Religion or Belief}, which includes provisions for gender equality. The declaration emphasizes the importance of protecting the rights of individuals, including women, and promotes the eradication of discrimination based on religion or belief. \newline

4. \textbf{International Day of Living Together in Peace}

\quad \textbf{Resolution 72/130} (adopted on 8 December 2017) designates 28 December as the International Day of Living Together in Peace. While not explicitly mentioning gender equality, this resolution underscores the importance of mutual respect and understanding, which are critical for gender equality in religious and spiritual contexts. \newline

5. \textbf{United Nations Alliance of Civilizations}

\quad \textbf{Resolution 69/312} (adopted on 6 July 2015) reaffirms the \textbf{United Nations Alliance of Civilizations}, which promotes dialogue among civilizations, cultures, religions, and beliefs. This initiative aligns with the goal of fostering gender equality by encouraging respectful interactions between diverse groups, including religious and spiritual communities. \newline

6. \textbf{Strategic Initiatives and Campaigns}  

\quad \textbf{UNESCO's Unite for Heritage Campaign} (launched in 2015) and the \textbf{Plan of Action to Safeguard Religious Sites} (prepared by the United Nations Alliance of Civilizations) highlight the importance of preserving cultural heritage and promoting tolerance. These initiatives indirectly support gender equality by ensuring that religious and spiritual practices are respected and accessible to all, including women and girls. \newline

7. \textbf{Combating Violent Extremism and Hate Speech}

\quad \textbf{Resolution 73/285} (adopted on 2 April 2019) calls for the elimination of violence based on religion or belief and emphasizes the role of interreligious and intercultural dialogue in countering hate speech. This resolution supports gender equality by addressing systemic issues that contribute to discrimination and violence, particularly against women and girls. 
\begin{minipage}[t]{\linewidth}

\end{minipage}
\\
\end{tabular}
}
\end{table}
\begin{table}[H]
\centering
\resizebox{\textwidth}{!}{
\begin{tabular}{p{\columnwidth}}

8. \textbf{International Day of Commemoration of the Victims of Acts of Violence Based on Religion or Belief}  

\quad \textbf{Resolution 73/296} (adopted on 28 May 2019) designates 22 August as the International Day of Commemoration of the Victims of Acts of Violence Based on Religion or Belief. This day highlights the need to address violence rooted in religious or spiritual discrimination, which disproportionately affects women and girls. \newline

9. \textbf{Global Migration and Cultural Diversity} 

\quad \textbf{Resolution 74/23} (adopted on 12 December 2019) acknowledges the growing importance of interreligious and intercultural dialogue in the context of migration. This resolution reinforces the idea that cultural diversity and the pursuit of cultural development are sources of mutual enrichment, which is essential for gender equality in a multicultural society.\newline 

10. \textbf{Role of Freedom of Expression in Combating Intolerance} 

\quad \textbf{Resolution 73/296} also reiterates the importance of the \textbf{right to freedom of opinion and expression} in strengthening democracy and combating religious intolerance. This principle is central to gender equality, as it ensures that women and other marginalized groups can express their views and participate in societal discourse without fear of discrimination. \newline

\textbf{Conclusion}:

\quad The UN resolutions collectively emphasize the intersection of religion, spirituality, and gender equality through the promotion of interfaith dialogue, the protection of human rights, and the prevention of discrimination. They recognize that religious and spiritual diversity must be respected and integrated into policies that advance gender equality, social justice, and peace. By fostering mutual understanding and respect, these resolutions aim to create inclusive societies where all individuals, regardless of their religious or spiritual beliefs, can thrive. \\
\hline
 \rowcolor{gray!20!white} \multicolumn{1}{c}{\textcolor{all-minilm}{\texttt{all-MiniLM-L6-v2}} for document retrieval and \textcolor{qwen3-tiny2}{\texttt{Qwen3-0.6B}} for answer generation} \\
\textbf{A}: The references to gender equality and religion/spirituality in UN resolutions are found in the following context:

1. \textbf{Mainstreaming a Gender Perspective}

\quad \textbf{Resolution adopted by the Economic and Social Council on 7 June 2023} (Agenda item 12). 
\begin{itemize}
    \item It reaffirms its agreement from 1997/2 and recalls previous resolutions (e.g., 2011/6, 2012/24, 2013/16, etc.).
    \item It emphasizes the importance of mainstreaming a gender perspective into all policies and programs. 
\end{itemize}
  
2. \textbf{Gender Equality and the Beijing Declaration and Platform for Action}
\begin{itemize}
    \item The UN has reiterated that gender mainstreaming is a critical strategy for achieving gender equality and empowering women and girls.
    \item The outcomes of the twenty-third special session and the implementation of the Programme of Action of the International Conference on Population and Development are highlighted.
\end{itemize}

3. \textbf{Global Commitments}  
\begin{itemize}
    \item The UN recognizes the importance of gender equality and empowerment through various summits and conferences, including the 2005 World Summit, 2011 World Summit, and the 2015 development agenda.
\end{itemize}

These references collectively emphasize the role of gender equality and the integration of religious/spiritual aspects into UN policies and initiatives.
\begin{minipage}[t]{\linewidth}
\end{minipage}
\\
\hline
\end{tabular}
}
\caption{Answers generated by SpiritRAG. The average response time is dependent on the system configuration. Our best system setup takes around 7 seconds to response.}
\end{table}

\section{Statistics on Domain-related Subjects in UN Resolution Documents}
\label{sec:subjects}
\begin{figure}[H]
    \centering
    \includegraphics[width=\columnwidth]{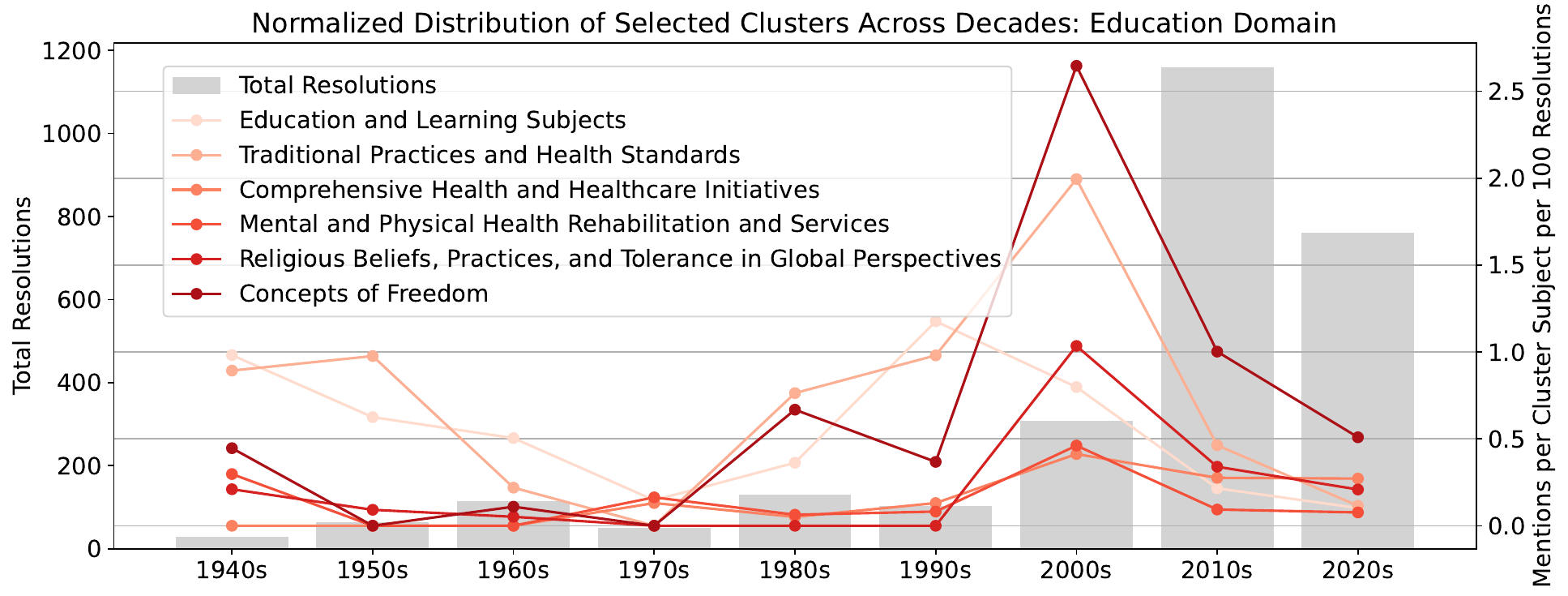}
    \caption{Temporal trends in the representation of selected subject clusters\textsuperscript{11} within the education domain. The gray bars indicate the total number of education-related resolutions in each decade. Colored lines show the normalized frequency of mentions for each cluster, calculated as the average number of subject mentions per cluster (per 100 resolutions), adjusted for the number of subjects in each cluster.}
    \label{fig:education-clusters}
\end{figure}

\begin{figure}[H]
    \centering
    \includegraphics[width=0.9\columnwidth]{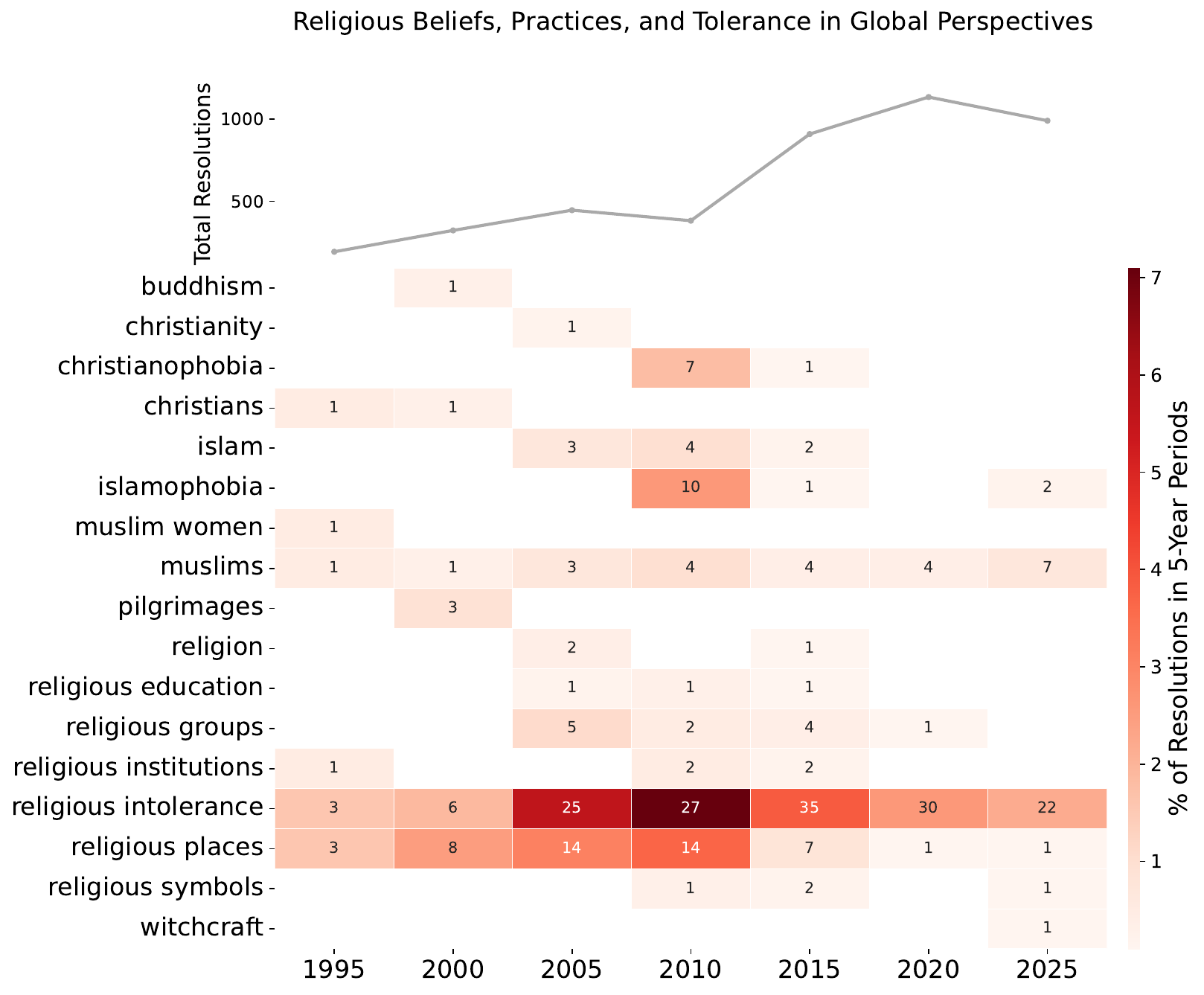}
    \caption{Distribution of the cluster \#133 subjects within the health domain across 5-year periods. Cell color reflects the proportion of resolutions in each period tagged with a given subject. Raw counts of subjects are overlaid as annotations. The line chart above shows the total number of health-related resolutions in each period.}
    \label{fig:religion-cluster}
\end{figure}

\footnotetext[11]{The clusters in Figures~\ref{fig:education-clusters} and~\ref{fig:religion-cluster} were created using Sentence-BERT (\texttt{all-MiniLM-L6-v2}) and agglomerative clustering from \texttt{sklearn} with the distance threshold 2. Cluster titles were assigned automatically using \texttt{GPT-4o}.}

\section{User Interface}
\label{sec:interface}
\begin{figure}[!htbp]
   \centering
   \begin{subfigure}[t]{0.9\textwidth}
        \centering
        \includegraphics[width=\textwidth]{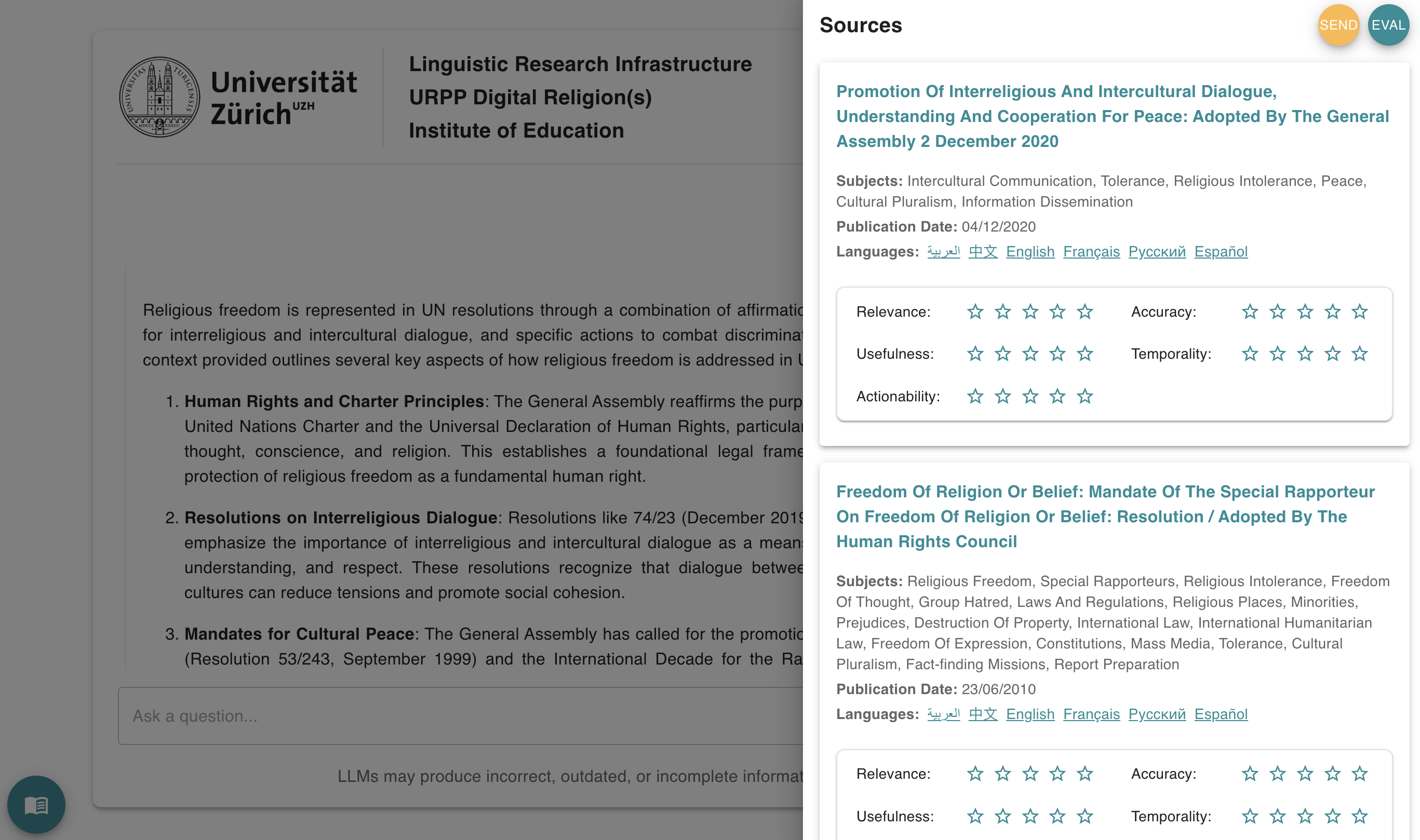}
        \caption{Retrieved documents.}
        \label{fig:right-panel}
    \end{subfigure}

    \vspace{5em}     
    \begin{subfigure}[t]{0.9\textwidth}
        \centering
        \includegraphics[width=\textwidth]{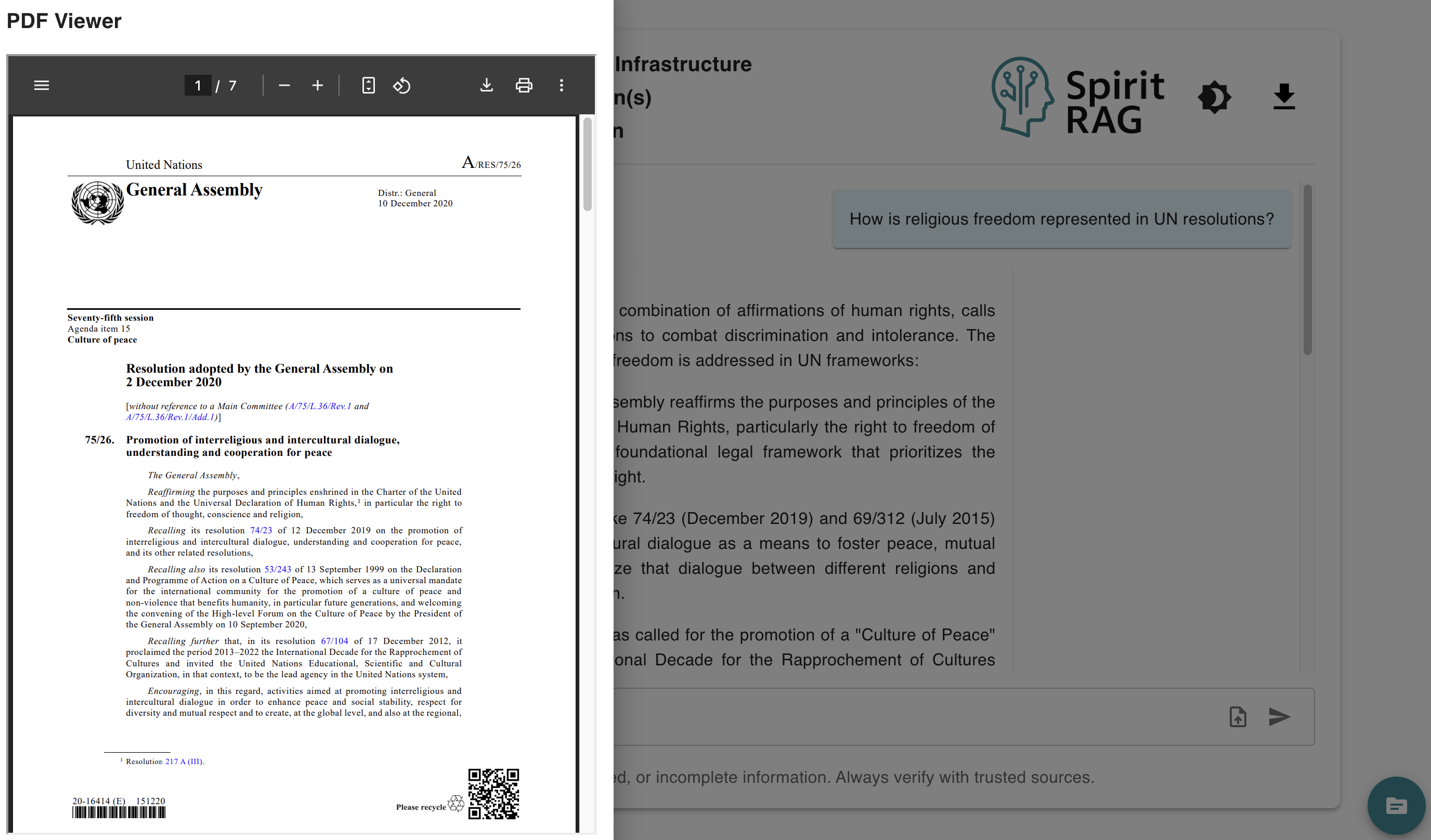}
        \caption{Document viewer.}
        \label{fig:left-panel}
    \end{subfigure}
    \caption{User interface of SpiritRAG. Left panel (viewing original PDF document) and right panel (viewing the list of the most relevant documents to the user question).}
    \label{fig:sidebar}
\end{figure}

\clearpage 
\let\clearpage\relax 

\end{document}